\documentclass{svproc}
\usepackage{url}
\usepackage{graphicx}
\usepackage{float}
\usepackage{subcaption}
\usepackage{orcidlink}

\begin{document}
\mainmatter              
\title{CHARACTER RECOGNITION OF NEPALI NUMBER PLATE}
\titlerunning{Character Recognition of Nepali Number Plate}  
%
\author{Satyasa Khadka \inst{1} \and Sandhya Baral \inst{1} \and
Sudip Tiwari \inst{1} \and Sharad Kumar Ghimire \inst{1}\orcidlink{0009-0009-3587-4529} \thanks{corresponding author} }
\authorrunning{S. Khadka et al.} 
%
\tocauthor{Satyasa Khadka, Sandhya Baral, Sudip Tiwari and Sharad Kumar Ghimire}
\institute{Pulchowk Campus, Institute of Engineering, Lalitpur, Nepal\inst{1}\\
\email{skghimire@ioe.edu.np} }

\maketitle              

\begin{abstract}
This paper presents a robust Automatic Number Plate Recognition (ANPR) system tailored for Nepali license plates written in Devanagari script. In this paper, a pipelined model was used that integrates YOLO-based models for license plate and character detection, followed by a CNN classifier trained on 34 Devanagari characters. Two publicly available data sets were used that incorporate diverse lighting, fonts, and structural variations. Data augmentation and additional training on embossed plates enhanced the generalizability of the model. The system achieved a recognition accuracy of up to 93\%, demonstrating strong performance under real-world conditions and providing a scalable solution for traffic management in Nepal.


\end{abstract}
\section{Introduction}
Automatic Number Plate Recognition (ANPR) has become an essential component of modern traffic surveillance and control systems. These systems integrate computer vision and machine learning algorithms to localize and recognize vehicle license plates from images. While significant progress has been made in ANPR technologies for standardized plate formats, adapting such systems to regional scripts and structures remains a challenge.

In the Nepali context, number plates contain Devanagari characters and exhibit high variability in terms of layout, font style, font size, and alignment. 
Plates may appear in one, two, or three rows, lack of standardization and, due to manual painting, lead to inconsistencies in spacing and character forms. These factors hinder the effectiveness of conventional ANPR pipelines, which often assume Latin-script-based, uniformly spaced characters.

Existing ANPR models trained on Latin character datasets are inadequate for recognizing Devanagari license plates due to script-specific complexities and positional character semantics. Moreover, many models assume fixed-length character sequences and uniform font styling, assumptions that do not hold in the Nepali setting. So, a customized system is needed that is suitable for Nepali license plates in Devanagari script.

The main objective of our study is to recognize the Nepali license plates in Devanagari script. To address this requirement, we prepared a robust ANPR system tailored for Nepali number plates using Convolutional Neural Networks (CNNs). Our approach includes plate detection, structure-aware character segmentation, and recognition using script-specific CNN models. The system is trained and validated on a custom-built dataset of Nepali plates, encompassing variations in ownership classes, row structures, and plate appearances.
\vspace{\baselineskip} 
\section{Literature Review}
Automatic Number Plate Recognition (ANPR) has been extensively studied, evolving from traditional machine learning approaches to deep learning-based pipelines. Early systems focused on hand-crafted features and classical classifiers, while recent methods leverage Convolutional Neural Networks (CNNs) to achieve higher robustness across variable conditions.

Pant et al.\cite{pant2015automatic} developed one of the first Nepali ANPR systems using Histogram of Oriented Gradients (HoG) features and a Support Vector Machine (SVM) classifier. Although effective on a limited dataset, the system struggled with variations in plate structure, achieving 75\% accuracy.

Recent work by Pandey et al.\cite{pandey2023cnn} applied CNN-based object detection with SSD MobileNet for plate localization and super-resolution techniques to enhance recognition. Character segmentation was handled using bounding boxes, resulting in 93\% recognition accuracy.

Dawadi et al.\cite{dawadi2023approach} proposed a comprehensive pipeline for Devanagari license plate detection, classification, and recognition using IWPOD-NET and nested Random Forest classifiers. Their approach addressed the complexity of Nepal’s multi-row, multilingual, and color-coded plates. Character recognition was performed using two dedicated CNN models trained on a custom dataset. Their system demonstrated high performance on both stationary and dynamic vehicle imagery, validating the model’s applicability in real-world scenarios.

\vspace{\baselineskip} 
\section{Methodology}
\begin{figure}[htbp]
\centering
\includegraphics[width=1.2\textwidth, height=0.2\textheight]{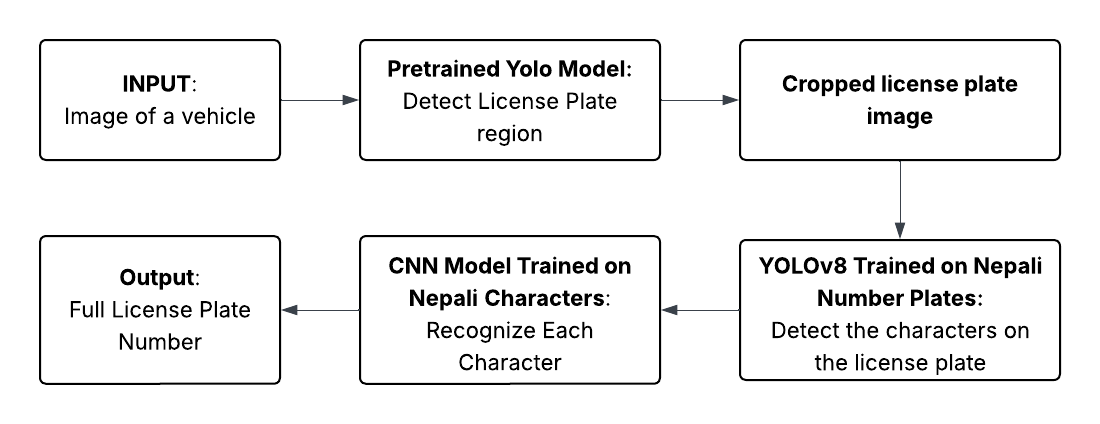}
\caption{System design of Nepali Number Plate Recognition System}
\label{fig:System design of Nepali Number Plate Recognition System}
\end{figure}
The general flow of the system is shown in figure \ref{fig:System design of Nepali Number Plate Recognition System} which is explained further below.
\subsection{Datasets Description}
This work utilizes two publicly available datasets from Kaggle, specifically tailored to the Nepali number plate recognition task:

\begin{itemize}
    \item \textbf{Nepali Vehicles Number Plate Dataset} \cite{vehicle_dataset}: This dataset contains approximately 2,500 vehicle images with annotated bounding boxes of license plates, captured under diverse environmental and lighting conditions. It serves as the primary source for training and evaluating plate detection and character segmentation models.
    
    \item \textbf{Nepali Number Plate Characters Dataset} \cite{character_dataset}: This dataset comprises 26,537 labeled images of the 34 unique Devanagari characters used on Nepali license plates. Data augmentation techniques such as rotation, scaling, and noise injection were applied to enhance model generalizability.
\end{itemize}

\subsection{License Plate Detection and Segmentation}
License plate detection was initially performed using a pre-trained YOLO model, chosen for its real-time detection capabilities and demonstrated efficacy across diverse operational contexts. The model predicts bounding boxes around license plates within vehicle images, enabling precise cropping of the plate region for subsequent processing.

For character-level segmentation, a YOLOv8 model was custom-trained on the Nepali license plate dataset. This approach facilitated accurate localization and extraction of individual characters, which were then forwarded to the recognition module. The adoption of YOLOv8 for segmentation substantially improved the system’s resilience to variations in character spacing, font styles, and background noise.

\subsection{Character Recognition}
Extracted character regions were classified using a Convolutional Neural Network (CNN) specifically trained to recognize the 34 Devanagari characters employed in Nepali license plates. The CNN architecture was meticulously optimized through grid search, guided by evaluation metrics such as classification accuracy and F1-score to ensure optimal model selection.

To address the challenge posed by embossed license plates—underrepresented in the primary datasets—a VGG-16-based CNN model was additionally trained using supplementary license plate images sourced from India and the United States. This augmentation enhances the system’s robustness, enabling reliable recognition across a wider spectrum of license plate types encountered in real-world scenarios.
\section{Results}
\subsection{Dataset} 
\begin{figure}[htbp]
\centering
\includegraphics[width=0.8\textwidth, height=0.3\textheight]{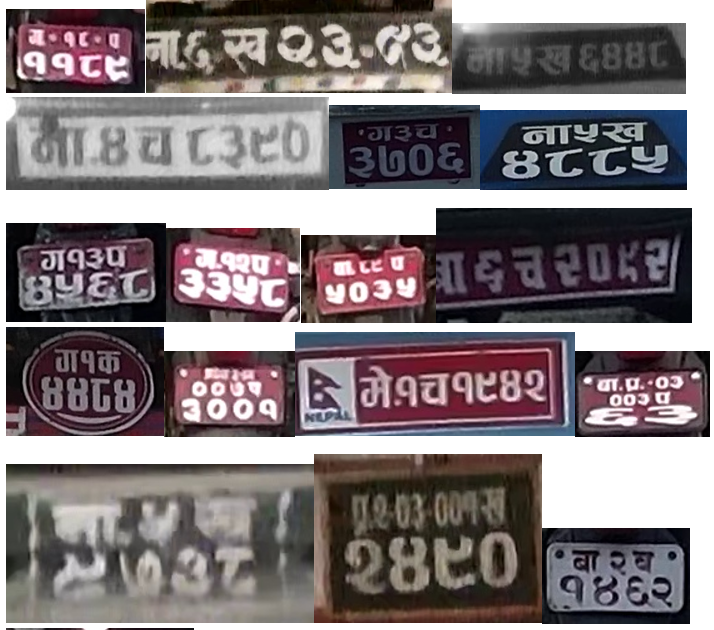}
\caption{Vehicles Dataset Collection}
\label{fig: Vehicles Dataset Collection}
\end{figure}

The license plate detection and character recognition components were trained using the datasets previously described in Section~3.1. Augmentation techniques such as brightness variation and perspective transformation were employed to enhance robustness under real-world conditions. Examples from license plate dataset is shown in figure~\ref{fig: Vehicles Dataset Collection}.

\begin{figure}[htbp]
\centering
\includegraphics[width=0.9\textwidth, height=0.3\textheight]{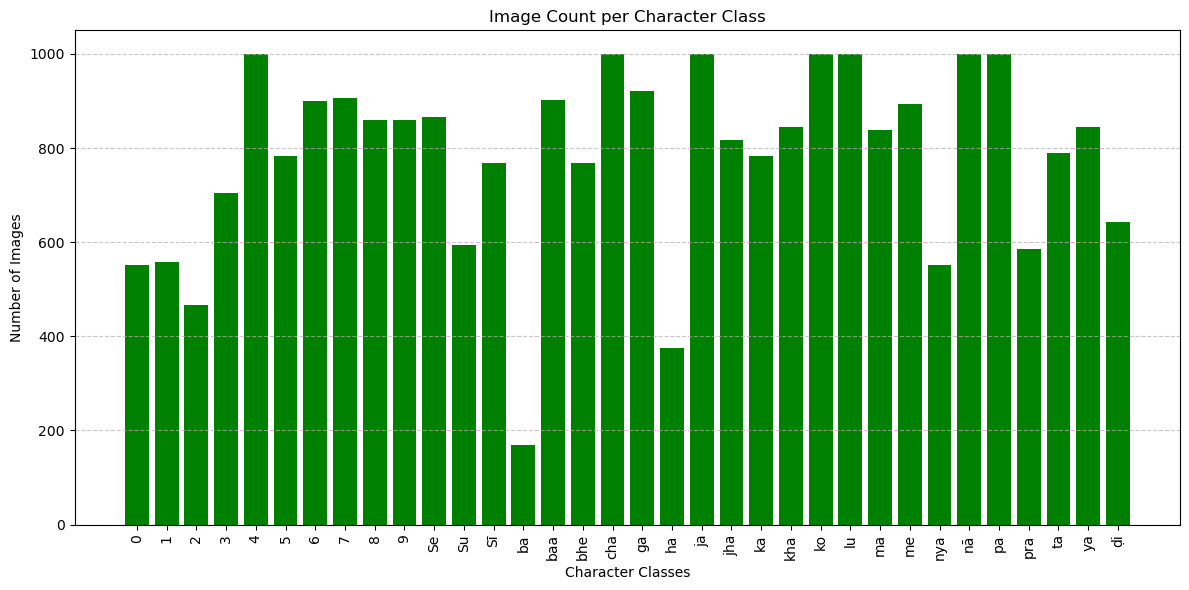 }
\caption{Character Dataset Distribution}
\label{fig: Character Dataset Distribution}
\end{figure}

The character recognition model was trained on the 34 unique Devanagari characters used in Nepali license plates, as described in Section~3.1. The dataset, comprising 26,537 labeled samples, was partitioned into training, validation, and test sets. Figure~\ref{fig: Character Dataset Distribution} illustrates the distribution of character classes. Data augementation strategies applied during training have improved the generalizability of the model and addressed class imbalance, contributing to overall recognition accuracy.

\subsection{ Performance Evaluation }
With the help of a pre-trained YOLO model, we have successfully detected the license plate from the image of a vehicle. The detected license plate was then fed into our trained YOLO model, which generated bounding boxes around the characters identified on the license plate.  

\begin{figure}[H]
\centering
\includegraphics[width=0.8\textwidth, height=0.3\textheight]{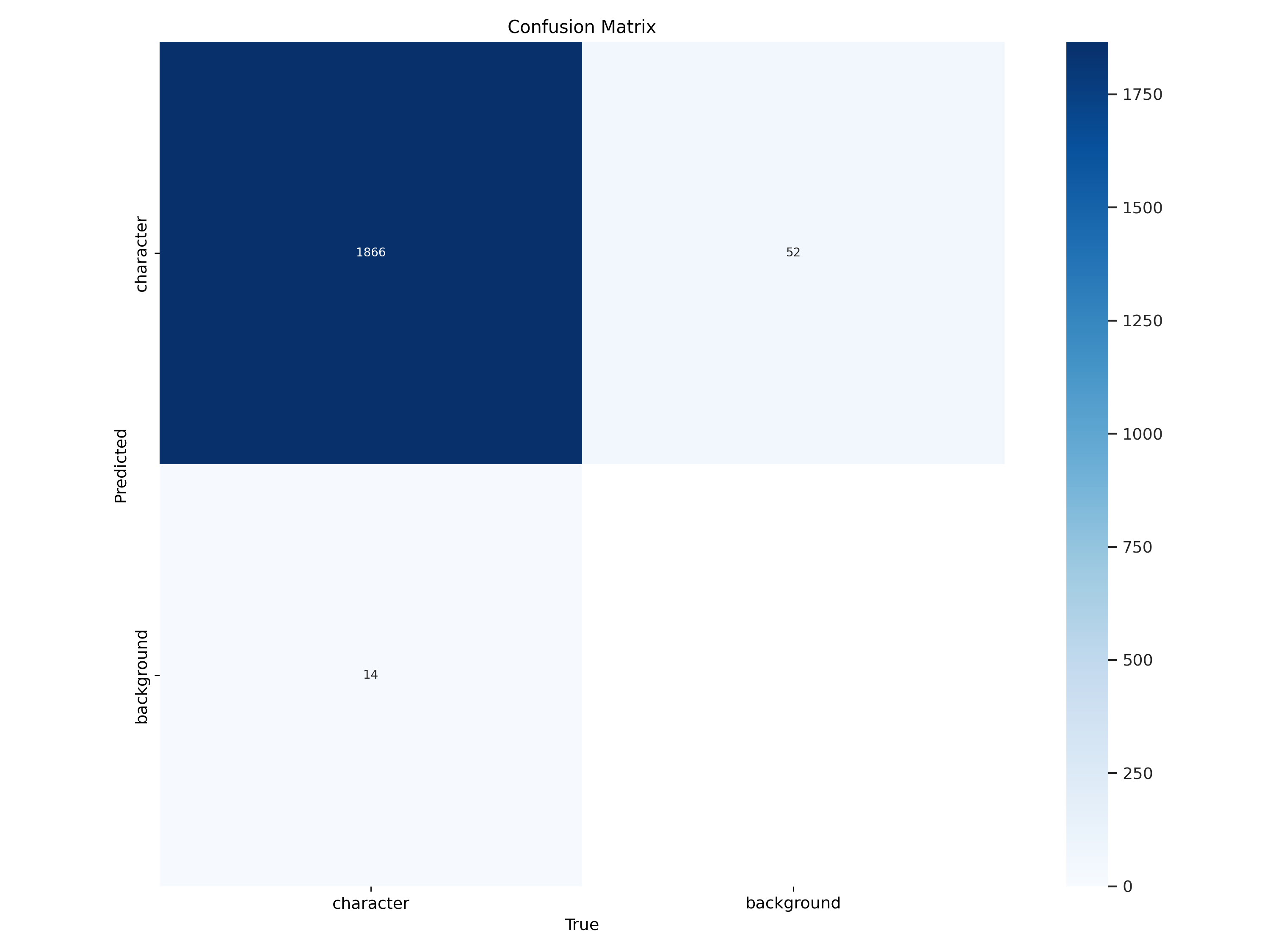 }
\caption{Confusion matrix of Trained YOLO Model}
\label{fig:YOLO_confusion matrix}
\end{figure}

Figure \ref{fig:YOLO_confusion matrix} shows the confusion matrix of our trained YOLO model.

\begin{figure}[htbp]

    \begin{subfigure}{0.8\linewidth}

    \centering
    \includegraphics[width=\linewidth]{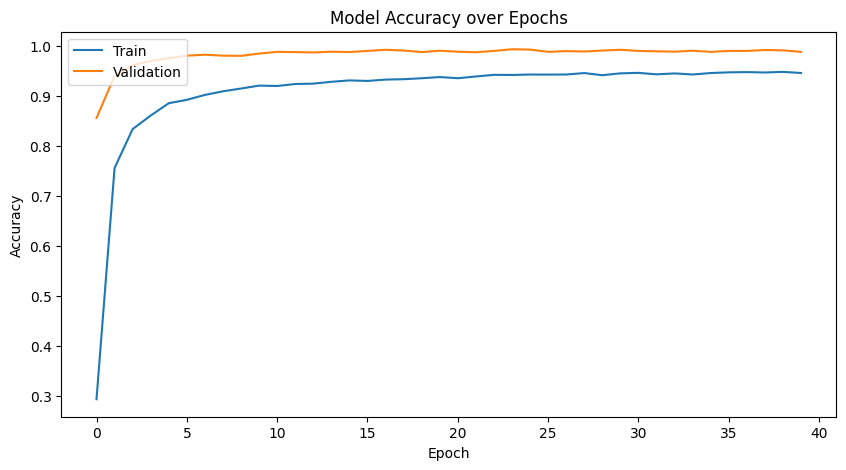}
    \caption{Training VS Validation accuracy}
    \label{fig:Accuracy plot}
    \end{subfigure}

    \begin{subfigure}{0.8\linewidth}

    \centering
    \includegraphics[width=\linewidth]{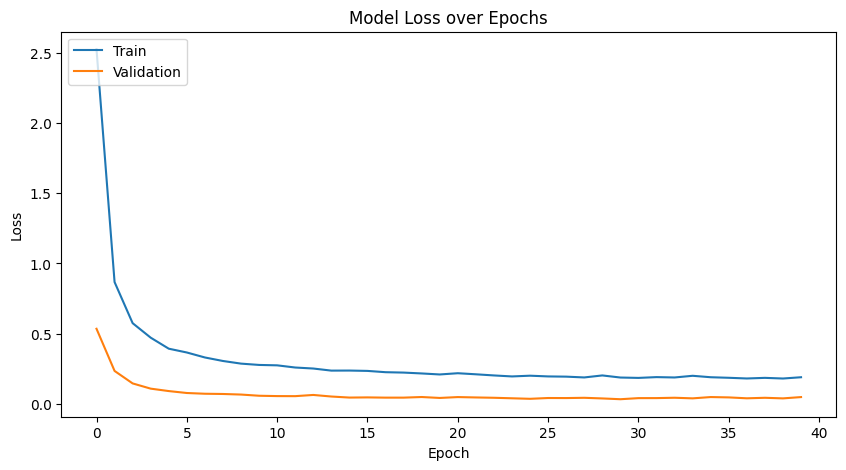}
    \caption{Training VS Validation loss}
    \label{fig:Loss plot}
    \end{subfigure}

    \caption{Training metrics}
    \label{fig:combined_metrics}

\end{figure}

The detected characters were cropped using the bounding boxes created by the YOLO model and subsequently fed into our CNN model for recognition. Figure~\ref{fig:combined_metrics} illustrates the accuracy and loss graphs of the CNN model during the training.

Finally, to demonstrate the methodology of this study, we developed a simple website. The website allows users to upload an image of a vehicle and it outputs the characters detected on the license plate using the pipeline model described above.

\begin{figure}[H]

    \centering
    \includegraphics[width=0.7\linewidth, height=0.35\textheight]{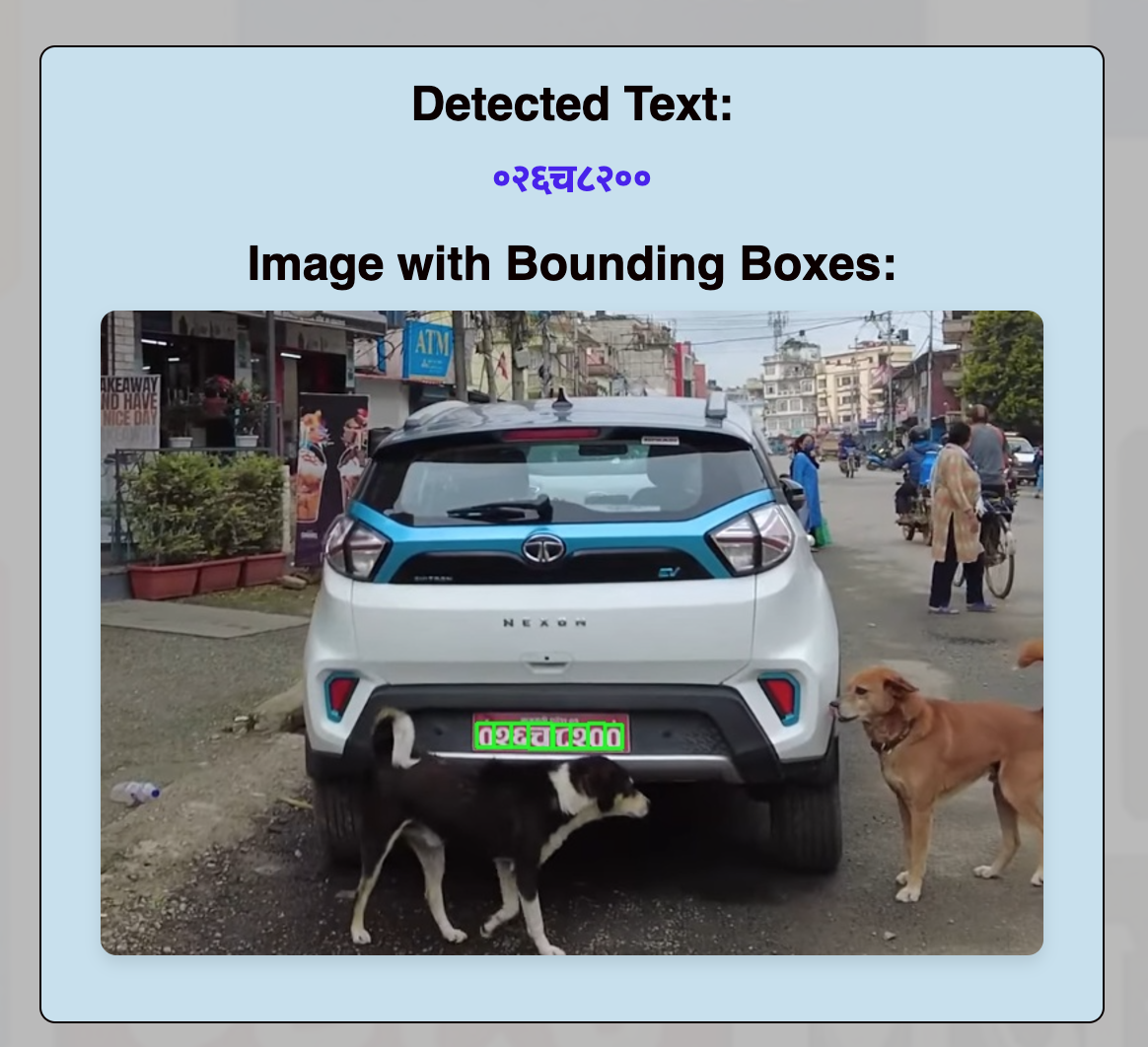}
    \caption{Final output}
    \label{fig:final output}
\end{figure}

The figure \ref{fig:final output} shows the final output of our website.

In addition to this, embossed number plates have become increasingly popular in Nepal in recent years. We have addressed the growing demand for automated recognition of Nepali number plates, by developing a model specifically to predict and recognize Nepali license plates. However, due to the lack of a dedicated dataset for Nepali embossed plates, we trained our model using images of license plates from India and the USA. The dataset comprises approximately 500 images, split into training and testing subsets. Our model achieved a final accuracy of around 93\% and a validation accuracy of 88\%. 

\begin{figure}[H]
\centering
\includegraphics[width=0.8\textwidth, height=0.28\textheight]{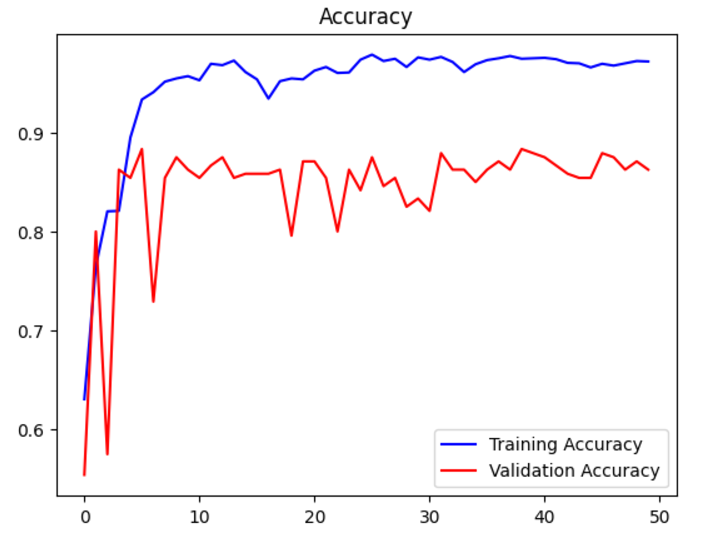}
\caption{Training vs Validation accuracy}
\label{fig:Training vs Validation accuracy}
\end{figure}
We implemented the VGG-16 model, a Convolutional Neural Network (CNN) architecture, for effective detection and classification of license plates.

\section{Conclusion}
This paper presented a robust and scalable system for automatic number plate recognition (ANPR) tailored to Nepali license plates written in the Devanagari script. By integrating YOLO-based models for plate and character detection with a CNN classifier trained on 34 Devanagari characters, the system achieved, strong real-world performance. Due to the lack of Nepali embossed samples, by the use of publicly available datasets, including a US embossed plate dataset, along with extensive data augmentation and supplemental training we became able to improve  the model’s generalization. Achieving up to 93\% recognition accuracy on diverse plate images, the proposed pipeline offers a significant step toward intelligent transportation systems and automated traffic surveillance in Nepal.

\section{Future Scope}
The proposed ANPR system shows promising results with an accuracy of 93\%, but still there is room for further improvement and expansion. A key future direction is the creation of a dedicated Nepali embossed plate dataset, as current models rely on foreign samples. Enhancing the model with advanced architectures like Vision Transformers and attention mechanisms can further boost recognition performance. Optimizing the system for edge devices using model compression techniques will support real-time deployment. Integration with intelligent transport systems and improving robustness under adverse conditions such as low light, motion blur, and occlusions are also important next steps. These advancements will help to build a more reliable and scalable ANPR solution for real-time deployment of Nepal's traffic management needs.

%
%

\end{document}